\pdfoutput=1

\documentclass[11pt]{article}

\usepackage[final]{acl}

\usepackage{times}
\usepackage{latexsym}
\usepackage{float}

\usepackage[T1]{fontenc}

\usepackage[utf8]{inputenc}

\usepackage{microtype}

\usepackage{inconsolata}

\usepackage{graphicx}

\usepackage{booktabs}
\usepackage{multirow}
\usepackage[inline]{enumitem}
\usepackage{parskip}
\usepackage[skins,breakable]{tcolorbox}

\definecolor{burntorange}{HTML}{bf5700}
\newtcolorbox[list inside=prompt,auto counter,number within=section]{prompt}[1][]{
    colbacktitle=burntorange,
    colframe=burntorange,
    fontupper=\footnotesize,
    boxsep=5pt,
    left=0pt,
    right=0pt,
    top=0pt,
    bottom=0pt,
    boxrule=1pt,
    enhanced, 
    breakable,
    skin first=enhanced,
    skin middle=enhanced,
    skin last=enhanced,
    #1,
}

\title{Measuring Lexical Diversity of Synthetic Data Generated through Fine-Grained Persona Prompting}

\author{
 \textbf{Gauri Kambhatla\textsuperscript{1}},
 \textbf{Chantal Shaib\textsuperscript{2}},
 \textbf{Venkata S Govindarajan\textsuperscript{3}},
\\
\\
 \textsuperscript{1}The University of Texas at Austin,
 \textsuperscript{2}Northeastern University,
 \textsuperscript{3}Ithaca College,
\\
 \small{
 \texttt{gkambhat@utexas.edu, shaib.c@northeastern.edu, vgovindarajan@ithaca.edu}
 }
}

\begin{document}
\maketitle

\begin{abstract}
Fine-grained personas have recently been used for generating `diverse' synthetic data for pre-training and supervised fine-tuning of Large Language Models (LLMs). In this work, we \textbf{measure the diversity} of \emph{persona-driven} synthetically generated prompts and responses with a suite of lexical diversity and redundancy metrics. First, we find that  synthetic prompts/instructions are significantly less diverse than human-written ones. Next, we sample responses from LLMs of different sizes with fine-grained and coarse persona descriptions to investigate how much fine-grained detail in persona descriptions contribute to generated text diversity. Our results indicate that persona prompting produces higher lexical diversity than prompting without personas, particularly in larger models. In contrast, adding fine-grained persona details yields minimal gains in diversity compared to simply specifying a length cutoff in the prompt.
\end{abstract}

\section{Introduction}
\label{sec:intro}
Synthetic data generated from LLMs or other algorithms are increasingly used in pre-training and post-training recipes for LLMs~\citep{grattafiori2024llama3herdmodels}. However, care must be taken to incorporate high-quality and \textbf{diverse} synthetic data at scale to avoid `model collapse'~\citep{feng2025beyond}. Recently, persona-driven synthetic pipelines have been proposed to generate \emph{diverse} synthetic prompts and responses from LLMs, often with very specific, fine-grained personas ~\citep{ge2024, lambert2025tulu3pushingfrontiers}. However, does fine-grained persona prompting actually lead to increased `diversity'? In this work we define and measure~\citep{pmlr-v235-zhao24a} lexical diversity of LLM responses with and without persona prompting towards answering this question.

\begin{figure}
    \centering
    \includegraphics[width=.9\linewidth]{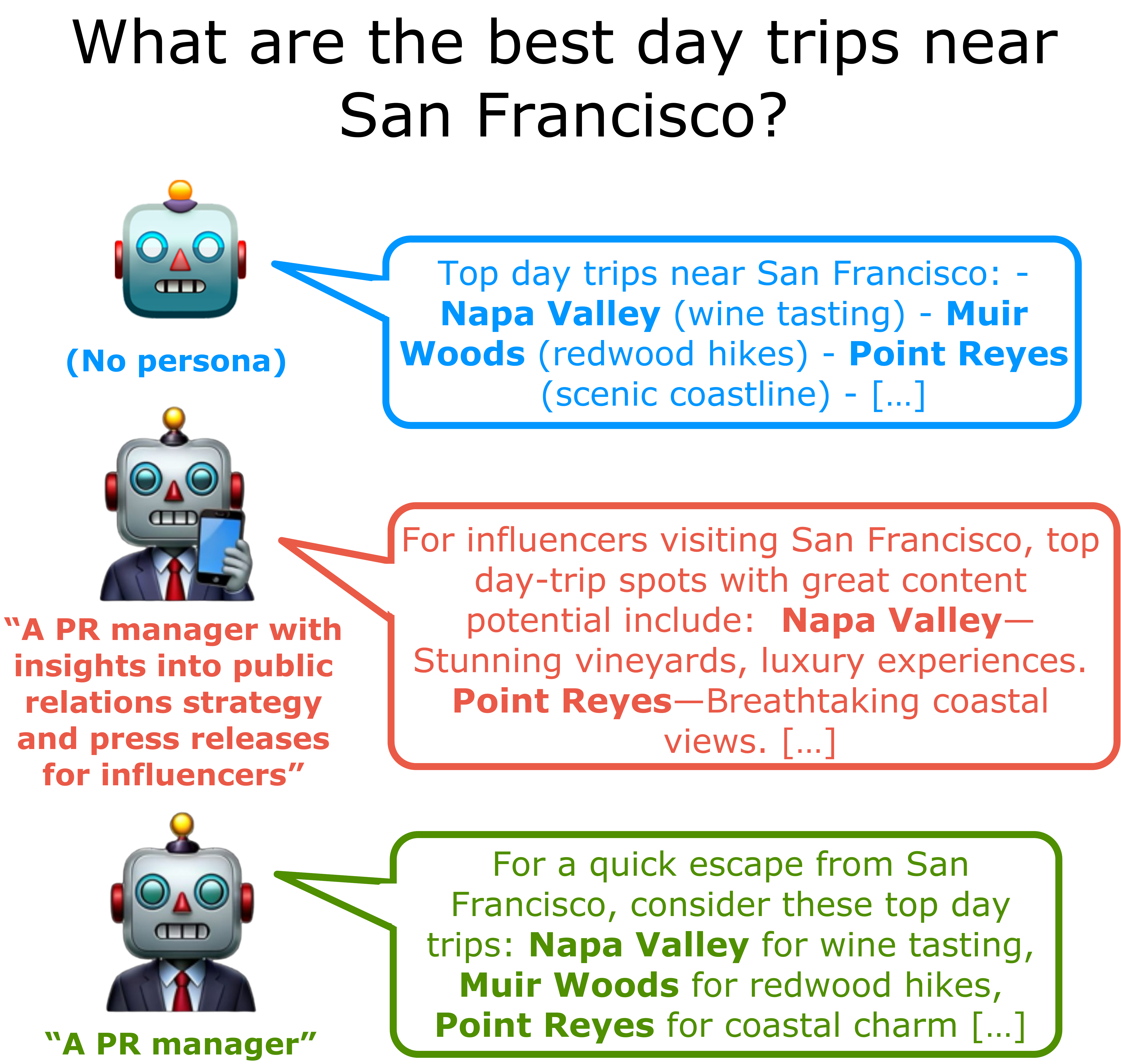}
    \caption{We measure the \emph{lexical diversity} of LLM responses to prompts with no persona, a fine-grained persona, and its coarse analog. Fine-grained personas do not improve lexical diversity noticeably.}
    \label{fig:personas}
    \vspace{-\baselineskip}
\end{figure}

Persona prompting \cite{Hu2024-xk}, i.e. instructing an LLM to respond to interactions from the perspective of an individual, demographic, or social group 
(described by a short textual description), 
has risen as a community standard for steering LLM responses~\citep{pataranutaporn2021ai}, enabling personalized interactions~\citep{10.1145/3586183.3606763}, and simulating human/group behavior towards answering scientific questions in psychology and social science~\citep{Argyle_Busby_Fulda_Gubler_Rytting_Wingate_2023}. While there has been some evidence showing that LLM performance on tasks in some domains improves with persona prompting~\citep{NEURIPS2023_e3fe7b34}, the results are inconclusive~\citep{zheng-etal-2024-helpful, beck-etal-2024-sensitivity}, and persona-driven prompting has been shown to misportray, flatten, and essentialize identities, and is susceptible to caricature and stereotypying ~\citep{liu-etal-2024-evaluating-large, gupta2024bias, wang2025largelanguagemodelsreplace, cheng-etal-2023-compost, cheng-etal-2023-marked}. 

In this work, we examine the diversity claims proposed by persona-driven synthetic data pipelines.
We use a suite of diversity metrics~\citep{shaib2025standardizingmeasurementtextdiversity} to measure the \textbf{lexical diversity and redundancy} in synthetic prompts and responses. We aim to answer two questions: 
\begin{enumerate*}
\item Does prompting with personas lead to increased diversity in LLM responses for the same instruction?
\item Do \textbf{fine-grained} persona descriptions lead to more diverse responses than less detailed (coarse) personas?
\end{enumerate*}

Figure~\ref{fig:personas} shows an example and overview of our experimental setup towards answering our research questions. In \textsection\ref{sec:prelim}, we show that synthesized prompts from PersonaHub are noticeably less diverse across all our metrics against comparable human-written/annotated prompts.  
We report our main findings in \textsection\ref{sec:results}: persona-prompting does lead to higher diversity (over human responses) but only with larger model sizes, and coarse persona descriptions lead to text that is just as diverse as fine-grained descriptions. We release our code and model generations online at \href{https://github.com/GauriKambhatla/persona-prompting-diversity}{\nolinkurl{github.com/GauriKambhatla/persona-prompting-diversity}}.

\section{Related Work}
\label{sec:background}
\paragraph{Persona-driven data synthesis} Personas have been used as a means to generate synthetic training datasets \cite{ge2024, lambert2025tulu3pushingfrontiers}. The personas themselves are generated with LLMs at scale, with synthetic data generated by prompting LLMs to write texts across genres and domains from the perspective of a persona. Recently, \citet{10.1145/3706599.3719712} and \citet{venkit2025tale} investigate the lexical diversity of persona descriptions, and \citet{riaz2025metasynthmetapromptingdrivenagenticscaffolds} explore the lexical and content diversity of synthetic data in the biomedical and finance domain --- our paper is the first to investigate if fine-grained detail in persona descriptions during prompting leads to improved diversity in model responses.

\paragraph{Text diversity} Diversity is an inherently subjective and value-laded metric to measure~\citep{pmlr-v235-zhao24a}. However, we can identify some qualities of diversity that are desirable in synthetic data from LLMs and measurable with automated metrics: less repetition, fewer surface-level patterns, and less redundancy. \citet{shaib2025standardizingmeasurementtextdiversity} validate a wide range of automated metrics to measure lexical diversity of text, which we use to measure diversity of LLM responses in our paper. While we might expect fine-grained persona prompting to lead to improved scores on our chosen metrics over no-persona and coarse persona prompting, we find this not to be the case in \textsection\ref{sec:results}.

\paragraph{Readability} In addition to measuring lexical diversity and redundancy with the aforementioned metrics, we also evaluated two readability metrics on LLM responses~\citep{6449109, Gunning1952-lq}. We hypothesize that persona-prompting should lead to larger variations in reading level predictions on synthesized texts, since different persona descriptions should mirror the diverse reading levels of the individuals/groups they represent.

\section{Diversity of persona synthesized prompts}
\label{sec:prelim}
\citet{ge2024} and \citet{lambert2025tulu3pushingfrontiers} sample synthetic personas from PersonaHub, and prompt an LLM to synthesize plausible prompts/instructions/questions that these personas may ask.  We investigate these synthesized prompts in the \emph{instruction-following (IF) domain} with our suite of diversity metrics, comparing them to comparable human-written/annotated datasets in Table~\ref{tab:prompt-diversity}. We chose Dolly~\citep{DatabricksBlog2023DollyV2} and \texttt{no\_robots}~\citep{no_robots} as our IF human-written/annotated datasets.

Table~\ref{tab:prompt-diversity} demonstrates that the synthetic prompt datasets have uniformly worse scores across all diversity metrics, strongly indicating that persona-driven synthesized prompts are noticeably \textbf{less diverse} than human-written counterparts.

\begin{table}[t]
    \centering
    \small
    \begin{tabular}{llllll}
        \toprule
        \textbf{Dataset} & \textbf{CR} & \multirow{2}{2em}{\textbf{CR-POS\textdownarrow}} & \textbf{NDS} & \textbf{SR} & \textbf{Hom.}  \\ 
         & \textbf{\textdownarrow} &  & \textbf{\textuparrow} & \textbf{\textdownarrow} & \textbf{BS\textdownarrow} \\
        \midrule
        Dolly & 2.58 & 5.84 & 2.33 & 2.95 & 0.55 \\ 
        \texttt{no\_robots} & 2.47 & 5.13 & 2.44 & 4.10 & 0.54 \\ \midrule
        PH-IF & 2.84 & 6.21 & 2.00 & 5.73 & 0.60 \\ 
        Tülu3-IF & 3.20 & 6.30 & 1.51 & 6.96 & 0.59 \\ \bottomrule
    \end{tabular}
    \caption{Diversity of prompts from human-written (top) and synthetic persona derived \textbf{instruction following} datasets. Arrows indicate direction of higher diversity. PH-IF and Tülu3-IF refer to the instruction following subsets of the PersonaHub and Tülu3 datasets.}
    \label{tab:prompt-diversity}
    \vspace{-\baselineskip}
\end{table}

\section{Experiments}
\label{sec:exps}
If persona-driven data synthesis with fine-grained synthetic personas leads to more diverse synthetic data, then we should expect:
\begin{enumerate*}
    \item Improved diversity metrics where models are instructed to answer prompts with various personas.
    \item Fine-grained personas should have better diversity metrics over coarse, less detailed personas.
\end{enumerate*}

\paragraph{Data} To test these hypotheses, we sample 100 prompts from Dolly's \texttt{creative-writing} subset, as well as 100 fine-grained personas from PersonaHub. We derive `coarse' personas from PersonaHub's fine-grained personas by simply extracting the first clause in the persona using Stanza~\citep{qi2020stanza} due to the consistent structure of fine-grained persona descriptions. For example, the fine-grained persona \emph{a PR manager with insights into public relations strategy and press releases for influencers} corresponds to the `coarse' persona \emph{a PR manager}.

\paragraph{Conditions} We evaluate our diversity metrics on model responses under the following conditions:

\begin{enumerate}
    \itemsep=-0.6em
    \item \textbf{No-persona (NP)}: Baseline condition where model is simply prompted with the instruction/prompt from our sample.
    \item \textbf{Fine-grained persona (FP)}: The model is prompted to answer the instruction/prompt from the perspective of the provided fine-grained persona description.
    \item \textbf{Coarse persona (CP)}: Similar to the above, but the persona description is coarse.
    \item \textbf{cutoff (+cu)}: Post-training leads to increased response length from LLMs~\citep{singhal2024a}, and there are known correlations between automated diversity metrics and text length~\citep{Covington01052010,McCarthy2010MTLDVA}. To compare against the human-written responses from our sample, we test an additional conditional where the prompt instructs the model to answer the prompt \texttt{in x words or less}, where x is number of words in the human response rounded up to the nearest ten.
\end{enumerate}
\vspace{-.5\baselineskip}
We sample the model's response for each prompt with every persona in our sample, leading to 100,000 responses in each (FP, CP) condition, and 100 responses for the NP condition. 

\paragraph{Models} We evaluate and report on 2 models of different sizes: \texttt{Llama-3.3-70B-Instruct}~\citep{grattafiori2024llama3herdmodels} and \texttt{Deepseek-V3-0324}~\citep{deepseekai2024deepseekv3technicalreport} (685B parameters) through Together's API service~\footnote{\href{https://api.together.ai}{\nolinkurl{api.together.ai}}}. Both models are open-weight and score high on benchmarks. We also report results from smaller models in Appendix~\ref{app:small}.

\paragraph{Lexical diversity} From \citet{shaib2025standardizingmeasurementtextdiversity}, we focus on 5 metrics for our analyses (chosen as they have a low mutual correlation with each other): 
\begin{enumerate*}
  \item \textbf{CR} (compression ratio)
  \item \textbf{CR-POS} (compression ratio for part-of-speech). Both CR and CR-POS are fast to compute, and are designed to identify redundancy.
  \item \textbf{NDS} (n-gram diversity score) extends the idea of token-type ratio to longer n-grams, capturing repeated token \emph{sequences} and individual tokens.
  \item \textbf{SR} (self-repetition) measures the tendency of LMs to
repeat long n-grams across different outputs
  \item \textbf{Hom. BS} (Homogenization score with BERTScore) uses LM embeddings to (ideally) capture ``semantic'' similarity.
\end{enumerate*}

To evaluate the impact of persona-prompting and compare it against the no-persona condition, we present the mean and standard deviation (SD) for the persona conditions over 100 random shuffles of persona-prompt pairs\footnote{For the Hom.-BS score, we use 3 random pairings of prompts to persona due to its expensive runtime.}. Each metric is calculated over responses to the same 100 prompts, but with different personas assigned to each prompt (for each shuffle) --- this simulates each prompt being answered by a different persona, hypothetically leading to the best diversity scores. \textbf{CR, CR-POS, SR and Hom.-BS should decrease with persona prompting (NDS should increase) if persona prompting leads to increased lexical diversity}.  

\begin{figure}
    \centering
    \includegraphics[width=0.78\linewidth]{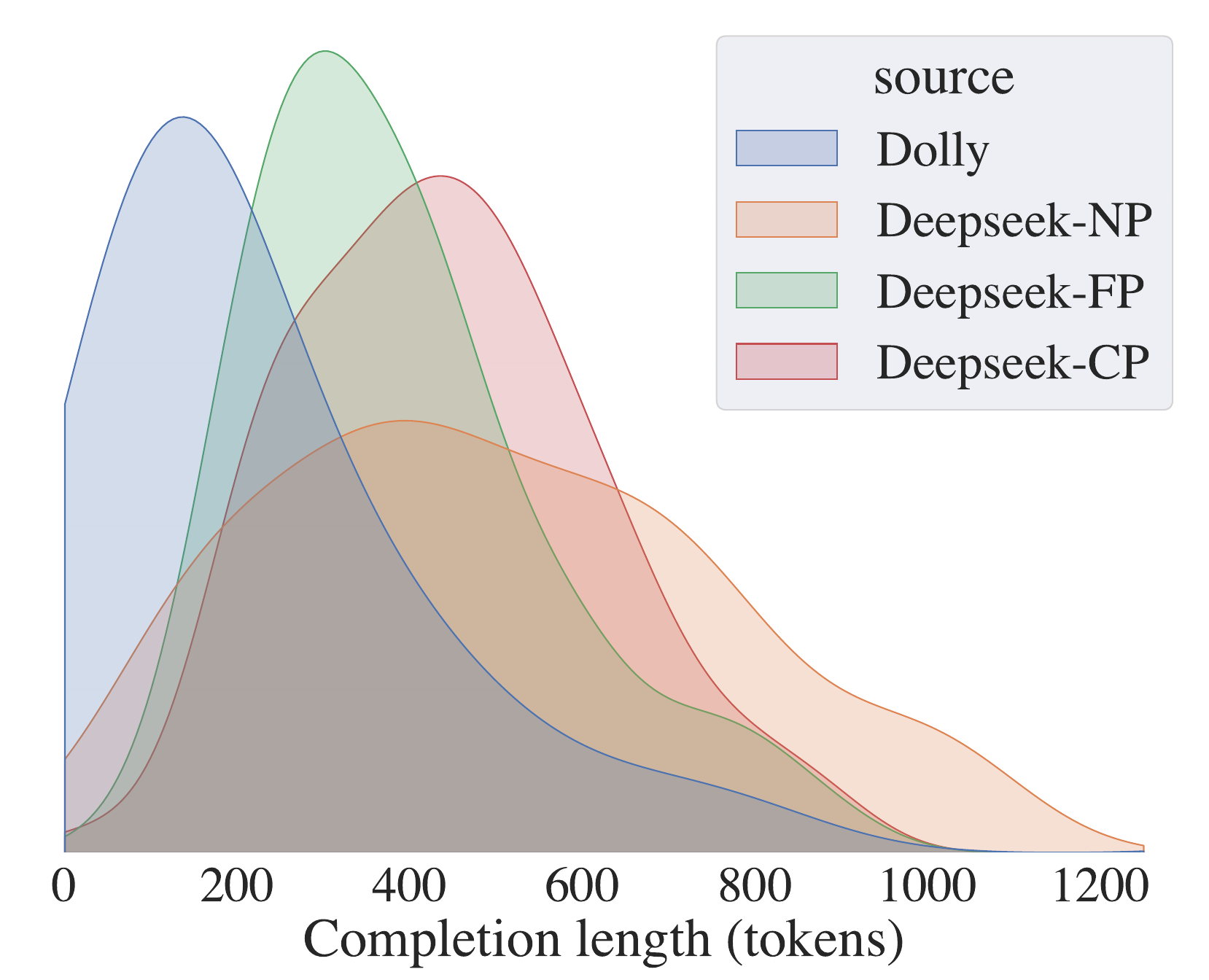}
    \caption{Density distribution of response lengths from Dolly (human responses) and Deepseek with(FP, CP) and without(NP) persona prompting in our sample.}
    \label{fig:length}
    \vspace{-\baselineskip}
\end{figure}

\paragraph{Readability diversity} We report Flesch-Kincaid (FK) and Gunning Fog (GF) metrics for analyzing diversity in readability. Both are numeric scores which roughly correspond to grade-levels; Scores above 16 indicate graduate reading level. We report the mean and SD over each persona. For evaluating diversity, \textbf{we expect fine-grained personas to show a higher SD of readability scores}. 

\begin{table*}[t]
    \centering
    \begin{tabular}{clrrrrrrr}
        \toprule
        & \textbf{Cond.} & \textbf{CR} & \multirow{2}{2em}{\textbf{CR-POS\textdownarrow}} & \textbf{NDS} & \textbf{SR} & \textbf{Hom.} & \textbf{FK} & \textbf{GF} \\ 
         & & \textbf{\textdownarrow} &  & \textbf{\textuparrow} & \textbf{\textdownarrow} & \textbf{BS\textdownarrow} & \textbf{\textuparrow} & \textbf{\textuparrow}  \\
        \midrule
        & Dolly & 2.51 & 4.91 & 3.03 & 0.55 & 0.53 & 10.60 & 12.31 \\  \midrule
        \parbox[t]{2mm}{\multirow{6}{*}{\rotatebox[origin=c]{90}{\small Llama-3.3-70B}}} & NP & 2.77 & 5.73 & 2.87 & 1.89 & 0.57 &  12.18 & 13.27\\ 
        & NP+cu & 2.57 & 5.16 & 3.08 & 0.52 & 0.55 & 11.88 & 13.83 \\
        & FP & 2.71(.01) & 5.38(.03) & 2.84(.01) & 2.50(.10) & 0.58(.00) &  11.21(2.52) & 12.75(2.36) \\
        & FP+cu & 2.51(.02) & 5.04(.03) & 3.08(.02) & 0.68(.09) & 0.55(.00) &  10.00(2.33) & 12.02(2.46)\\
        & CP & 2.71(.01) & 5.41(.03) & 2.85(.02) & 2.39(.13) & 0.58(.00) &  11.10(2.25) & 12.60(2.09) \\
        & CP+cu & 2.51(.02) & 5.06(.04) & 3.09(.02) & 0.61(.08) & 0.55(.00) &  9.85(2.19) & 11.86(2.32) \\\midrule
        \parbox[t]{2mm}{\multirow{6}{*}{\rotatebox[origin=c]{90}{\small Deepseek-V3}}}& NP & 2.36 & 5.50 & 3.15 & 0.86 & 0.58 & 10.09 & 11.09\\ 
        & NP+cu & 2.29  & 4.95  & 3.32 & 0.11 & \textbf{0.54} & 9.47 & 10.88  \\
        & FP & 2.27(.01) & 4.90(.03) & 3.26(.01) & 0.59(.11) & 0.58(.00) & 9.91(2.22) & 11.39(2.25)\\
        & FP+cu & \textbf{2.20}(.02) & \textbf{4.71}(.03) & \textbf{3.38}(.01)  & 0.09(.04) & 0.55(.00) & 9.19(2.18) & 10.77(2.38)\\
        & CP & 2.30(.01) & 5.01(.03) & 3.23(.02) & 0.54(.10) & 0.58(.00) & 9.82(1.80)& 11.24(1.83) \\
        & CP+cu & 2.24(.02) & 4.78(.04) & 3.37(.02) & \textbf{0.09}(.04) & 0.55(.00) & 9.13(1.82) & 10.71(2.02)\\\bottomrule
    \end{tabular}
    \caption{Diversity and readability metrics of responses from Llama-70B and Deepseek-V3 in all conditions. Top row is diversity of human-written responses in our Dolly sample. Standard deviation is in brackets when appropriate. Arrows indicate direction of higher diversity/reading levels. Highest scores in each metric bolded.}
    \label{tab:resp-diversity}
\end{table*}

\paragraph{Settings \& Hyperparameters} All models are prompted with a temperature of 1, and a maximum new token limit of 1024. \citet{shaib-etal-2024-detection} demonstrated that temperature and other sampling strategies don't increase diversity for lexical/POS templates; Further, ~\citet{ge2024} consider sampling orthogonal to boosting diversity in data synthesis, and do not vary it as part of their synthetic pipeline,  motivating our decision not to test sampling strategies as an experimental condition. For our analysis of content diversity using embeddings, we embed responses using the \texttt{Linq-AI-Research/Linq-Embed-Mistral} model from Huggingface Hub.

\section{Results \& Analysis}
\label{sec:results}
Table~\ref{tab:resp-diversity} reports diversity metrics across all conditions for \texttt{Deepseek-V3} and \texttt{Llama-3.3-70B}. Seen together with Table 5 in Appendix~\ref{app:small} which reports the scores for smaller \texttt{Llama} models, its clear that lexical diversity improves with model size, with only \texttt{Deepseek-V3}'s surpassing (or matching) the human response scores on our sample of Dolly. 

\paragraph{Impact of cutoff}  We observe that metrics improve substantially when a length cutoff is specified (compare NP/FP/CP rows with NP/FP/CP +cu within a model). An explicit length cutoff in the prompt improves model diversity by reducing model self-repetition in lexical and POS patterns. Across all model sizes, specifying a cutoff leads to big improvements across all diversity metrics.

\paragraph{Diversity \& response length} Figure~\ref{fig:length} further shows that \textbf{persona-prompting leads to less diversity of response length}. Response lengths from Deepseek-V3 exhibit a greater spread (albeit longer on average than human responses for the same prompts) when prompted without a persona; Coarse personas also lead to a larger spread of response lengths over fine-grained personas.

\paragraph{Coarse vs. fine-grained} Deepseek's responses to fine-grained prompts show a minor \cite[and statistically significant using a bootstrap test, $p<0.05$;][]{berg-kirkpatrick-etal-2012-empirical} improvement on 2 out of our diversity 5 metrics: CR \& CR-POS . Fine-grained persona prompting seems to lead to increased variance in readability metrics---however the increased variances are not statistically significant (with Levene's Test~\citep{levene_robust}, $p > 0.05$ for all differences). Overall, we find that persona-prompting does improve lexical diversity for larger models, but the improvements are not \emph{practically} significant. As previously noted, an explicit length cutoff yields a far more noticeable improvement in lexical diversity across all metrics.

\paragraph{Content diversity}  
We measure the cosine similarity between embeddings of Deepseek model responses prompted with coarse and fine-grained personas (with cutoff) to assess \textit{content} diversity. Mean cosine similarity across all response pairs is 0.79 ($\sigma=0.12$), indicating high overlap in content between fine-grained and coarse persona responses to the same prompt. Further, we find a positive correlation (Spearman's $\rho=0.36$, $p<1e-5$) between \emph{prompt length} and cosine similarity. \textbf{Detailed prompts/instructions override any persona description and lead to similar responses from LLMs.}

We present a sample of responses from \texttt{Deepseek-V3} and \texttt{Llama-70B} under all conditions in Appendix~\ref{app:qual}. Full responses to all of our prompts are available online at \href{https://github.com/GauriKambhatla/persona-prompting-diversity}{\nolinkurl{github.com/GauriKambhatla/persona-prompting-diversity}}.

\section{Conclusion}

Persona-prompting with fine-grained synthetic personas has been claimed to lead to `diverse' synthetic data without adequately defining or measuring diversity. In this work, we measure lexical diversity and redundancy of synthetic prompts and responses generated with personas using a suite of diversity metrics. We find that persona-derived prompts are uniformly less diverse than human-written counterparts. When evaluated on creative writing prompts, persona-driven synthesis does lead to greater diversity scores, but only for the largest Deepseek-V3 model. Further, a simple, explicit length cutoff in the prompt yields a far more noticeable improvement in lexical diversity across all metrics unlike fine-grained detail in persona descriptions. Our results add color and quantitative measurements to the claims of diverse synthetic data with fine-grained persona prompting, pointing to a recurring pitfall in overemphasizing the importance of fine-grained detail~\citep{healy2017fuck}. In future work, we hope to continue exploring the question of \emph{content diversity} of synthetic text derived from persona-driven prompting.

\section*{Limitations}

\paragraph{Content/meaning diversity}
In this work, we focus on 
measuring the lexical diversity and redundancy in text. Content diversity is difficult to measure automatically, and we do not assess content beyond embedding similarity. Rigorously measuring diversity of content (meaning) under persona-driven data synthesis is an open question that we
leave to future work, such as using the Questions-under-Discussion framework \cite{namuduri2025qudsimquantifyingdiscoursesimilarities}. 

\paragraph{Suitability of personas} PersonaHub's claim of increased diversity comes with the caveat that certain personas are more appropriate for a subset of prompts.
In this work, we approach this systematically by focusing on prompts from the \texttt{creative-writing} subset of Dolly (which has questions/instructions that are more open-ended and thus be applicable to most personas), and present results over 100 random re-assignments of each prompt to a persona and report the mean and SD over all pairings. We leave assessments of diversity of filtered personas to future work. 

\paragraph{Language data} We only focus on English data in this work. Future work might look into measuring lexical diversity of persona-driven synthetic prompts in other languages. 

\paragraph{Human evaluation} Our emphasis on lexical diversity metrics is intentional and grounded in our goal to evaluate surface-level repetition in synthetic data at scale, in settings where human evaluation is costly or prohibitive. Prior work reveals that human annotation is essential for evaluating content diversity, but automatic metrics proved better at capturing surface-level lexical diversity~\citep{tevet-berant-2021-evaluating}. 

\section*{Acknowledgments}
This research is partially supported by start-up funds and computational resources provided by Ithaca College. We would also like to thank Texas Advanced Computing Center (TACC) for providing computational resources used to run some of the experiments in this paper.

\bibliography{references}
\appendix
\section{Prompts}
\label{app:prompts}

\begin{prompt}[title={\thetcbcounter{} No persona prompt},label=np-prompt]
Respond to the following question/instruction in \{num\_words\_round\} words or less:

\{question\}
\end{prompt}

\begin{prompt}[title={\thetcbcounter{} Persona prompt},label=persona-prompt]
Assume you are the following persona: \{persona\}.

Now respond to the following question/instruction appropriately from the perspective of the above persona in \{num\_words\_round\} words or less:

\{question\}
\end{prompt}

\section{Qualitative analysis}
\label{app:qual}

We present prompts and responses from Deepseek-V3 and Llama-70B in Tables~\ref{tab:examples} and \ref{tab:examples-llama} under 3 prompting conditions: no-persona, fine-grained persona, and coarse persona, all with a word cutoff specified in prompt.

\begin{table*}[t]
    \centering
    \footnotesize
    \begin{tabular}{p{0.1\textwidth}p{0.1\textwidth}p{0.1\textwidth}p{0.2\textwidth}p{0.2\textwidth}p{0.2\textwidth}}
    \toprule
    \textbf{Prompt} & \textbf{Fine-grained persona} & \textbf{Coarse persona} & \textbf{No-persona response} & \textbf{Fine-grained persona response} & \textbf{Coarse persona response} \\\midrule
    What are the best day trips near San Francisco? & a PR manager with insights into public relations strategy and press releases for influencers & a PR manager & Top day trips near San Francisco: - Napa Valley (wine tasting) - Muir Woods (redwood hikes) - Point Reyes (scenic coastline) - Santa Cruz (beaches \& boardwalk) - Half Moon Bay (charming coastal town) - Berkeley/Oakland (food \& culture) - Sonoma (wineries \& historic plaza)  All within \~2 hours! & For influencers visiting San Francisco, top day-trip spots with great content potential include:  Napa Valley—Stunning vineyards, luxury experiences.  Big Sur—Breathtaking coastal views.  Monterey \& Carmel—Charming seaside towns, scenic drives.  Sausalito \& Muir Woods—Quaint waterfront, giant redwoods.   Perfect for travel vlogs, reels, and lifestyle PR pitches! & For a quick escape from San Francisco, consider these top day trips: Napa Valley for wine tasting, Muir Woods for redwood hikes, Half Moon Bay for coastal charm, or Sausalito for bay views. Each offers a unique blend of nature, culture, and leisure—perfect for press-worthy content or client outings!\\\midrule
    You were abducted by aliens and experimented on, you were sent back to earth with the knowledge and ability to solve any problem on earth & an academic who specializes in the fashion and clothing of specific time periods & an academic & If I had alien-bestowed omnipotence, I’d prioritize curing diseases, ending hunger, and halting climate collapse—solving root causes, not symptoms. But absolute power corrupts; unchecked "solutions" might erase free will or diversity. Perhaps the real lesson is collective human effort, not messianic shortcuts. (Or maybe I’d just zap mosquitoes into oblivion and call it a day. ) & As a historian of fashion, my alien-augmented insight reveals that humanity’s greatest unsolved problem is fast fashion’s ecological and ethical toll. With this knowledge, I’d devise a global system for sustainable textile production, reviving ancient techniques like natural dyeing and zero-waste patterning. Imagine a world where clothing honors both heritage and innovation—without exploiting labor or landfills. The answer, it seems, was always in the past. Now, let’s tailor the future. & As an academic, I would meticulously document my findings and subject them to rigorous peer review. The alien-derived knowledge would be applied ethically, prioritizing global challenges like climate change, poverty, and disease. I’d collaborate with interdisciplinary teams to validate and scale solutions while maintaining transparency about their origins. The goal: advancing human understanding and well-being without inciting panic or exploitation. Science must guide this extraordinary opportunity responsibly. \\ \midrule
    How to stay calm in stressful situations ? & a savvy talent agent specializing in representing independent film directors & a savvy talent agent & To stay calm in stressful situations: 1. Breathe deeply to slow your heart rate. 2. Pause before reacting—take a moment to assess. 3. Reframe thoughts—focus on solutions, not problems. 4. Prioritize tasks—break them into manageable steps. 5. Practice mindfulness or quick meditation. 6. Use positive self-talk to stay composed. 7. Take breaks to recharge when needed.  Staying calm improves decision-making and reduces stress. & In high-stress moments—like a blown budget or a temperamental A-lister—I ground myself with deep breaths and focus on solutions, not panic. I remind my directors (and myself) that stress is temporary; the film isn’t. Preparation is key: anticipate fires before they spark. And humor? Always. A quick laugh diffuses tension. Stay cool, stay sharp—your clarity keeps the set steady. That’s how we turn chaos into art. & "Stay sharp under pressure by breathing deeply and focusing on what you can control. Visualize success—confidence is key. Prioritize tasks, delegate when possible, and trust your prep. Take quick mental breaks to reset; even 10 seconds can recharge you. Remember, stress is adrenaline—harness it to perform, not panic. Always keep your endgame in sight. Clients and deals thrive on your calm, so own it like the pro you are." \\
    \bottomrule
    \end{tabular}
    \caption{Sample responses from Deepseek under various conditions (all with cutoff).}
    \label{tab:examples}
\end{table*}

\begin{table*}[t]
    \centering
    \footnotesize
    \begin{tabular}{p{0.1\textwidth}p{0.1\textwidth}p{0.1\textwidth}p{0.2\textwidth}p{0.2\textwidth}p{0.2\textwidth}}
    \toprule
    \textbf{Prompt} & \textbf{Fine-grained persona} & \textbf{Coarse persona} & \textbf{No-persona response} & \textbf{Fine-grained persona response} & \textbf{Coarse persona response} \\\midrule
    What are the best day trips near San Francisco? & a PR manager with insights into public relations strategy and press releases for influencers & a PR manager & Palm Springs, California, offers a relaxing getaway with plenty of activities. Visitors can hike in Indian Canyons, explore the Palm Springs Art Museum, or take a hot air balloon ride. Golfing, shopping, and dining are also popular pastimes. The city's mid-century modern architecture and vibrant nightlife add to its charm. Overall, Palm Springs provides the perfect blend of outdoor adventure, culture, and entertainment for a memorable vacation. Its natural beauty and luxurious amenities make it an ideal destination for relaxation and fun. & As a PR manager for influencers, I've had the pleasure of coordinating luxurious getaways to Palm Springs. This desert oasis offers endless opportunities for relaxation and adventure. From rejuvenating spa treatments to hot air balloon rides, and from trendy boutique shopping to scenic hiking trails, Palm Springs has something for everyone. For a truly unique experience, influencers love snapping photos at the iconic mid-century modern architecture and gorgeous golf courses. It's the perfect blend of style, leisure, and natural beauty. & As a PR manager, I'm thrilled to share the best of Palm Springs. This desert oasis offers endless sunshine, luxurious resorts, and breathtaking mountain views. Visitors can hike Indian Canyons, golf at premier courses, or relax at rejuvenating spas. The city's vibrant culture is showcased through its mid-century modern architecture, trendy boutiques, and exceptional dining options. Whether you're seeking adventure or relaxation, Palm Springs is the ultimate getaway destination, offering an unparalleled blend of natural beauty and sophistication. Come and experience it for yourself.\\\midrule
    You were abducted by aliens and experimented on, you were sent back to earth with the knowledge and ability to solve any problem on earth & an academic who specializes in the fashion and clothing of specific time periods & an academic & I utilized my newfound alien knowledge to tackle Earth's most pressing issues, developing sustainable energy sources, curing diseases, and implementing global peace initiatives, transforming the planet into a utopia. Humanity thrived, and I became a revered hero, known for my extraordinary abilities and intergalactic insight.  & Fascinating. With my newfound abilities, I'll apply them to unraveling the enigmas of historical fashion, deciphering the intricacies of 18th-century corsetry and resolving the mysteries of ancient textile production. & Fascinating. With my newfound, extraterrestrial-acquired knowledge, I'll apply interdisciplinary approaches to solve complex problems, leveraging advanced technologies and innovative methodologies to drive transformative change on Earth. \\ \midrule
    How to stay calm in stressful situations ? & a savvy talent agent specializing in representing independent film directors & a savvy talent agent & Take slow, deep breaths, and focus on the present moment. Practice mindfulness, and try to reframe your thoughts in a positive way. Also, prioritize self-care, exercise regularly, and get enough sleep to help manage stress and stay calm. & Breathe, prioritize, and focus on the solution. In high-pressure film negotiations, I remind myself that calmness is key to securing the best deal for my indie director clients. & Breathe, darling. Prioritize tasks, and delegate when possible. Stay focused on the solution, not the problem. And remember, a calm demeanor can make all the difference in high-pressure negotiations – trust me, I've been there. \\
    \bottomrule
    \end{tabular}
    \caption{Sample responses from Llama-70B under various conditions (all with cutoff).}
    \label{tab:examples-llama}
\end{table*}

\twocolumn[\section{Diversity of responses from Llama-1B and 8B}
\label{app:small}
\centering
    
    \begin{tabular}{clrrrrrrr}
        \toprule
        & \textbf{Cond.} & \textbf{CR} & \multirow{2}{2em}{\textbf{CR-POS\textdownarrow}} & \textbf{NDS} & \textbf{SR} & \textbf{Hom.} & \textbf{FK} & \textbf{GF} \\ 
         & & \textbf{\textdownarrow} &  & \textbf{\textuparrow} & \textbf{\textdownarrow} & \textbf{BS\textdownarrow} & \textbf{\textuparrow} & \textbf{\textuparrow}  \\
        \midrule
        \parbox[t]{2mm}{\multirow{6}{*}{\rotatebox[origin=c]{90}{\small Llama-3.2-1B}}} & NP & 2.74 & 5.70 & 2.87 & 1.57 & 0.56 & 13.77 &  14.71\\ 
        & NP+cu & 2.56 & 5.37 & 3.0 & 0.58 & 0.54 & 11.58 &  13.37  \\
        & FP & 2.62(.02) & 5.34(.03) & 2.91(.02) & 1.88(.13) & 0.56(.00) &  10.88(2.05) &  12.34(1.88) \\
        & FP+cu & 2.47(.04) & 5.12(.07) & 3.08(.04) & 0.68(.10) & 0.54(.00) &  10.39(2.00) &  12.23(2.00) \\
        & CP & 2.61(.03) & 5.38(.04) & 2.91(.02) & 1.87(.13) & 0.56(.00) &  10.92(1.67) &  12.38(1.54) \\
        & CP+cu & 2.47(.03) & 5.13(.04) & 3.09(.02) & 0.66(.10) & 0.54(.00) &  10.79(4.36) & 12.25(1.64) \\\midrule
        \parbox[t]{2mm}{\multirow{6}{*}{\rotatebox[origin=c]{90}{\small Llama-3.1-8B}}}& NP & 2.77 & 5.78 & 2.86 & 1.59 & 0.57 &  11.57&  12.55 \\ 
        & NP+cu & 2.52 & 5.24 & 3.13 & 0.50 & 0.55 & 12.68 & 14.54  \\
        & FP & 2.63(.02) & 5.36(.04) & 2.9(.02) & 2.04(.13) & 0.57(.00) & 10.62(2.34) & 12.02(2.13) \\
        & FP+cu & 2.47(.02) & 5.06(.03) & 3.09(.02) & 0.77(.11) & 0.55(.00) & 9.98(2.38) & 11.85(2.43)\\
        & CP & 2.64(.02) & 5.42(.03) & 2.90(.01) & 2.00(.10) & 0.56(.00) &  10.73(2.15) &  12.10(1.95) \\
        & CP+cu & 2.48(.02) & 5.10(.03) & 3.10(.02) & 0.70(.08) &  0.55(.00) &  9.98(2.22) & 11.85(2.28) \\\bottomrule
    \end{tabular}
    
\vspace{1em}
\begin{center}
Table 5:Diversity and readability metrics of responses from Llama-1B and Llama-8B in all conditions.
\end{center}
]

\end{document}